# A Comprehensive Study on Various Statistical Techniques for Prediction of Movie Success

*Manav Agarwal, Shreya Venugopal, Rishab Kashyap, R Bharathi

Department of CSE, PES University, Bangalore, India

## ABSTRACT

*The film industry is one of the most popular entertainment industries and one of the biggest markets for business. Among the contributing factors to this would be the success of a movie in terms of its popularity as well as its box office performance. Hence, we create a comprehensive comparison between the various machine learning models to predict the rate of success of a movie. The effectiveness of these models along with their statistical significance is studied to conclude which of these models is the best predictor. Some insights regarding factors that affect the success of the movies are also found. The models studied include some Regression models, Machine Learning models, a Time Series model and a Neural Network with the Neural Network being the best performing model with an accuracy of about 86%. Additionally, as part of the testing data for the movies released in 2020 are analysed.*

## 1. INTRODUCTION

One of the most important contributing factors to the entertainment industry are movies, which is one of the highest revenue-generating businesses from a commercial perspective [1]. A majority of the population love to watch a variety of movies, and their choices are determined based on the various factors that contribute to the type of movie such as the genre of the movie. Most people thus look into the ratings of a given movie before they proceed to watch it to identify it as a movie which is worth watching. These ratings come from a variety of sources, some of which include popular websites such as Rotten Tomatoes, IMDb and many more. Thus, the analysis involves the study of these user ratings and the other factors that affect the movie and this enables the prediction of whether a movie is truly a successful one or not.

## 2. LITERATURE SURVEY

We begin our survey by selecting an appropriate dataset [2], since it proves to be the foundation for our project. We also apply data mining [3-4] to extract movies released in 2020. The steps that follow involve creating a study on the various features that complement a movie's success, and performing a thorough study on them as done by Abidi et. al [1]. These attributes are fed into a variety of regression [5] Machine Learning techniques [6-7] and Neural Network [8] models. We thus create a comparative study between them as done by Dhir and Raj [9] who have created a comparison specifically among Machine Learning models. Looking into the variety of models provided in the various papers [10-11], we predict the success rate [4] using a subset of these with respect to the attributes chosen earlier. Furthermore, for each model, proving their statistical significance using a variety of tests [12-13] proves to be the highlight of our paper. In comparison with the accuracies provided in the papers [14] and their respective models [9], our models end up with a similar accuracy of 86% as shown through our papers.

## 3. PROPOSED METHODOLOGY

Similar to the methods mentioned above, through our paper we perform similar analytics on the IMDb dataset and create a comparison between the various models used [9], [11], and try to find an appropriate model as shown in Figure 1.

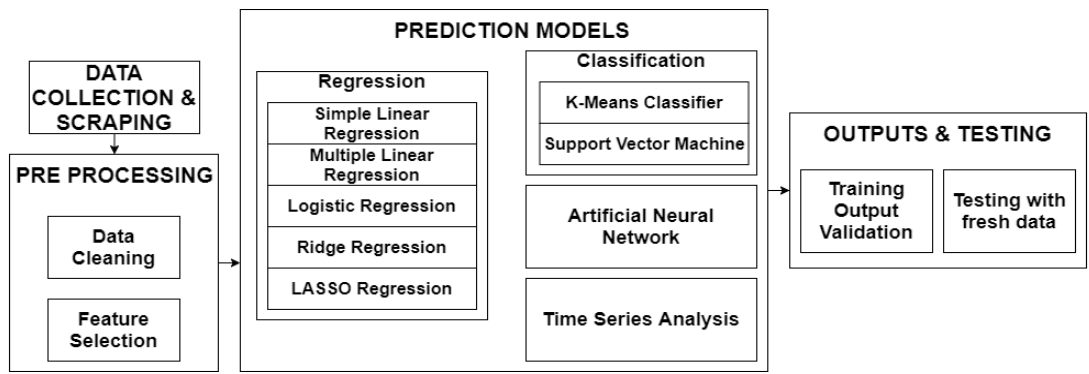

Figure 1. Proposed Methodology

## 4. DATASET DESCRIPTION

As mentioned above, we have considered the IMDb dataset [2] for our project. Some of the major attributes include the movie rating itself, followed by the genre, top voter ratings, total votes, duration of the movie and the release date. The minor attributes are cleaned in the pre-processing stage. We consider a total of 81274 movies for this process, split into training and testing according to the year of release in the steps that follow. To add on to the testing dataset, we have mined the official IMDb website for all the movies released in the year 2020, consisting of the same attributes to maintain uniformity.

## 5. PRE-PROCESSING

Traditional pre-processing methods such as removing null values, outliers and other basic steps have been applied. Apart from these, another pre-processing step used is the MultiLabelBinarizer [15] which is a method used to encode attributes that can belong to more than one category such as genre. Genres belonging to a movie are marked with the value 1, and the rest of the genres are marked with 0. This process is repeated for all movies until we have a list of numeric values depicting the genres that belong to the movie, and those that do not. We took all movies from 1990 to 2015 as the training dataset and all movies after 2015 as our validation dataset. The dataset contains an attribute called the "metascore" which ranges from 0 to 100 and is used as a measure to depict the success of a movie. A higher metascore implies that the movie was more successful. Hence this is used as the dependent variable for the complete study. To classify a movie as a success or a failure, the metascore value is divided into classes or bins, representing movies that are a hit or a flop or mediocre [16]. The predicted values can be classified under this partitioning.

## 6. REGRESSION METHODS

### 6.1. Simple Linear Regression

As we all know, Simple Linear Regression (SLR) is the most basic regression form. Since it involves only one independent variable against the metascore the most suitable attribute is selected. To do so, we use the Variance Inflation Factor (VIF) method to show the attributes with their multicollinearity with respect to each other. The highest value is considered to be the best prediction value, and in this case, we obtain the attribute to be the top1000_voters_ratings. We hence use this attribute in the SLR model and train the model with the existing values. The VIF ranges from values 0 in the case of "budget" to around 0.55 in the case of "top1000_voters_rating" as depicted in Figure 2.

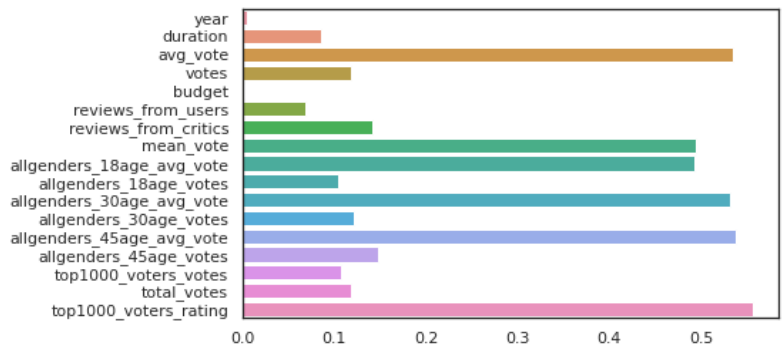

Figure 2. Diagram to show the VIF values for all attributes

Once we have our trained model, we use the testing dataset to predict the future values, and then make a comparison between the true and predicted values. The accuracy of the model is calculated using a confusion matrix. On completion of this, we proceed to perform tests on the model to prove its statistical significance.

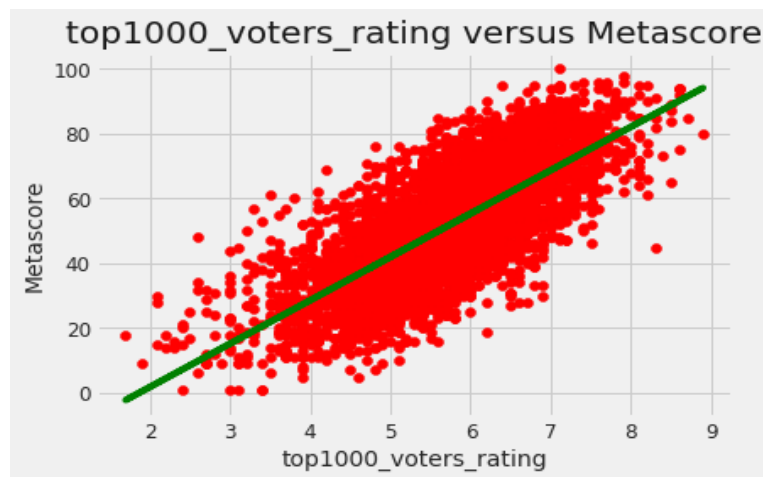

Figure 3. SLR model

We then test the same on our 2020 dataset, taking the "avg_vote" attribute i.e. the average rating of a movie instead of the "top100_voters_ratings" attribute in the previously trained model and get an accuracy of 0.5833. The model has already been tested for its significance and hence we do not test it again for the 2020 dataset.

## 6.2. Multiple Linear Regression

Multiple Linear regression (MLR) performs a process almost identical to SLR, but with multiple independent variables used to predict the same target variable. Similar to the procedure above, the target variable is predicted using the top1000_voters_rating attribute along with fifteen other attributes. Here, twelve of these attributes come under the MultiLabelBinarized values of the genre. The VIF test performed on them shows us the initial attributes that can be considered to plot the model. Once again, we perform the OLS tests and several hypothesis tests such as Jarque Bera test and Lagrange's Multiplier test for attribute selection to show the significance of the values. The model that we get gives us an accuracy of 0.7116. An example of the MLR model in three dimensions using two independent variables is shown in Figure 4. We perform the same analysis using our 2020 dataset, taking the attributes avg_vote and duration for plotting our MLR model, and result with an accuracy of 0.608.

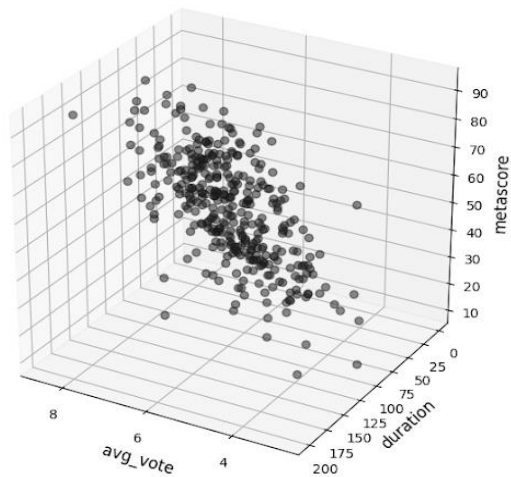

Figure 4. MLR model.

## 6.3. Logistic Regression

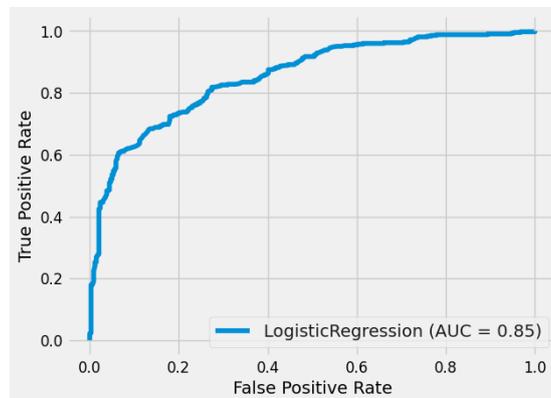

Figure 5. ROC-AUC curve for the logistic regression model

Logistic Regression is a non-linear regression and a statistical technique for finding the existence of a relationship between a qualitative and a quantitative dependent variable and several independent variables or explanatory variables [13]. The deviation we take from our originally defined steps here is, we divide our metascore into two domains, a successful or an unsuccessful movie based on the metascore value. This model results in an accuracy of 0.76. We plot an ROC-AUC curve as shown in Figure 5 using the confusion matrix that we have obtained. The same way we perform the Logistic analysis on our 2020 dataset to get an accuracy of 0.6833.

## 6.4. Regularization Techniques

Regularization is required to penalize certain features, and we have two regression methods for the same, namely the Ridge, and the LASSO regression models. Ridge regression is a technique used when the data suffers from high multicollinearity [12]. It uses the L2 regularization method to perform this process. The regularization parameter for Ridge Regression was 1150. The accuracy we get from this model comes up to 0.74. Plotting the same as in Fig. 6 using our 2020 dataset we thus get an accuracy of 0.61 where it can be inferred that the model is somewhat accurate however tests are conducted to validate the same.

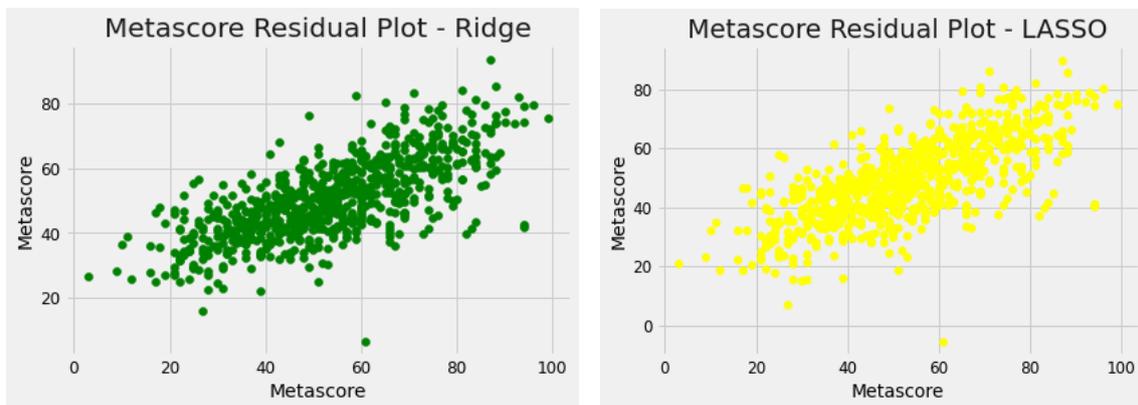

Figure 6. Ridge and LASSO regression model residual plot

As we did for Ridge, we follow the exact same procedure in LASSO regression, which along with L2 regularization, uses central tendency to penalize. The regularization parameter for LASSO was 0.145. For the training and testing process, calculating the statistical values as well as obtaining an accuracy of 0.72. Plotting the same using our 2020 dataset we thus get an accuracy of 0.59.

## 7. CLASSIFICATION METHODS

The classification models that we have used follow the unsupervised learning algorithm, rather they use self-learning techniques where there is no training data. The two classification models that we use here are the Support Vector Machine and the K-Means classifier.

## 7.1. K-Means Classifier

K-Means is a popular clustering technique and can also be used as a classifier where each cluster is considered as a class. Using the K-Means classifier in this case would involve making the value of K as 3 in this case for each of the metascore bins [12]. The accuracy of the K-Means model is 0.5. Statistical tests such as the silhouette test for a better model is performed. We plot as shown in Figure 7 the same for our 2020 dataset using our previously trained model and hence get an accuracy of 0.42.

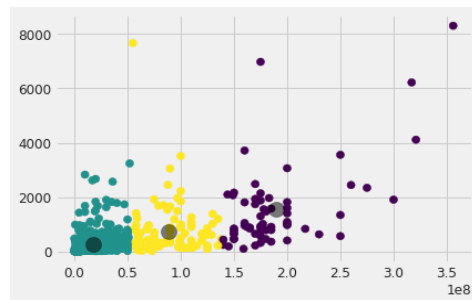

Figure 7. K-Means plot to show the final centroid positions

## 7.2. Support Vector Machine

Instead of the conventional binary Support Vector Machine (SVM) the one that classifies into 3 categories with two hyperplanes of separation is used. This is because the data is divided into 3 categories and it is such that they somewhat fall in a sequence. We can see that the model thus results in an accuracy of 0.71. The trained model is then used against the 2020 testing dataset to calculate the same and thus results in an accuracy of 0.62.

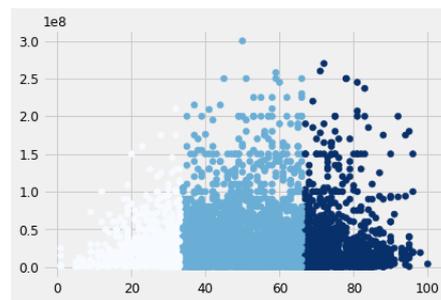

Figure 8. SVM models for our initial dataset

## 8. TIME SERIES ANALYSIS

We begin by defining the method of Forecasting, which is one of the most important and frequently used applications in analytics, which focuses on the prediction of future values based on the present and past values. We say that the variable is forecasted into future values to perform analysis on patterns such as trend, seasonality and so on. We make use of the SARIMAX model for our analysis. We first establish all the necessary parameters to plot our time series graphs. We then initialize the various attributes that will be used to determine the forecasted values for metascore. The time-wise analysis is based on the release date of the movie.

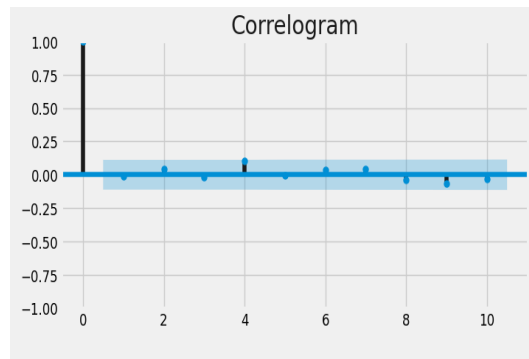

Figure 9. Correlogram Plot

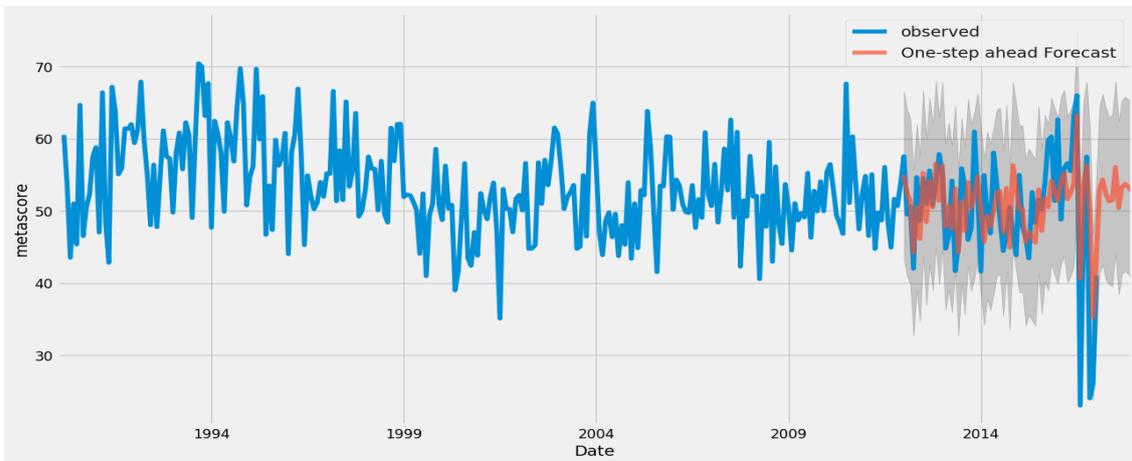

Figure 10. Residual plot with Forecasted values

On plotting the metascore value against the date_published attribute to get the time-series graph for the entire dataset, and then plotting the decomposition of the data into the trend, seasonality as well as the residual variations we obtain Figure 9 and Figure 10. To prove the stationarity of the data we use the Augmented Dickey-Fuller Test which revealed that the data is stationary. The presence of seasonality in the plot leads us to use the SARIMAX model. Initially we define the function to execute all possibilities for SARIMAX from which one model is selected based on its AIC value. We can see that the model gives us a roughly accurate result for the forecast. We plot the same for our 2020 dataset, using the SARIMAX model we used before to get similar outputs as in Figure 11.

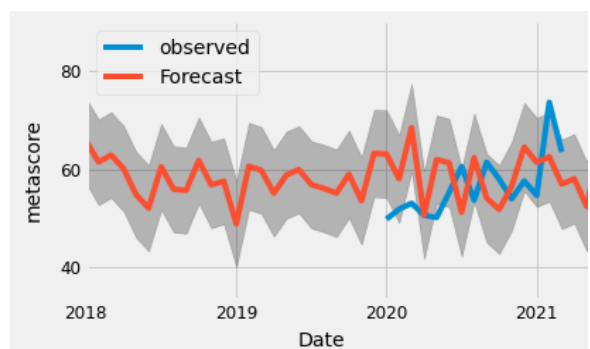

Figure 11. Plot of output using the model against the actual values.

## 9. ARTIFICIAL NEURAL NETWORK

In the Artificial Neural Network model, we have proposed, we give the inputs based on the statistical significance they hold from the tests we have performed in the previous cases. We obtain a final result of 86.16% accuracy from the model as well as minimal loss incurred through the process. The loss is also displayed in Figure 12. We have also tested this with the 2020 dataset obtained resulting in a 88.056 % accuracy.

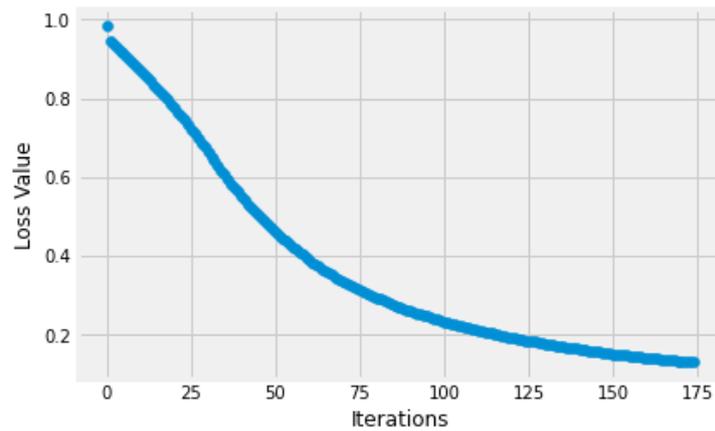

Figure 12. Loss Curve for the ANN

## 10. RESULTS AND INTERPRETATION OF VALUES

### 10.1. Regression

Wald's test for Logistic Regression proved the statistical significance of the model in Table 1.

Table 1. Wald's Test for Logistic Regression

| X | Chi^2 | p-value > Chi^2 | df constraint |
|---|---|---|---|
| const | 1062.614071 | 4.411953e-233 | 1 |
| top1000_voters_rating | 1104.475166 | 3.517505e-242 | 1 |
| Action | 53.033523 | 3.279040e-13 | 1 |
| Crime | 22.853796 | 1.748038e-06 | 1 |
| Drama | 24.430425 | 7.704234e-07 | 1 |
| Fantasy | 16.520951 | 4.811547e-05 | 1 |
| Mystery | 25.968786 | 3.469824e-07 | 1 |
| Romance | 3.914710 | 4.786528e-02 | 1 |
| Sport | 7.101317 | 7.702733e-03 | 1 |
| Thriller | 9.629585 | 1.914678e-03 | 1 |
| War | 4.919469 | 2.655568e-02 | 1 |

Hence the analysis shifts to the other regression models that are compared in Table 2. The Durbin Watson Test is a hypothesis test that checks for autocorrelation between error terms, the null hypothesis indicating that there is no autocorrelation. As a rule of thumb, a Durbin-Watson statistic close to 2 implies no autocorrelation. This is the case for all our regression models since it has Durbin-Watson statistic values ranging from 1.7163 as in the case of LASSO regression to 1.963 for Multiple Linear Regression. In terms of R2 values and its variants Multiple Linear Regression seems to be showing the highest value. This is supported by the F statistic or ANOVA (Analysis of Variance) which is a hypothesis test whose null hypothesis states that all regression coefficients of the model should be zero. Hence it is a test that ensures the overall regression of the model. In Multiple Linear Regression the F-statistic is high unlike that for Ridge and Lasso Regression indicating that the results are not significant. However, the F-statistic value is significantly higher for SLR which seems to give it along with MLR more credit compared to the other Regressions in the ranking of models as its R2 value was only marginally lower. The Jarque-Bera test is a goodness-of-fit test of whether sample data has the skewness and kurtosis matching a normal distribution. Lagrange Multiplier test (LM test also known as Breusch Godfrey test) is a hypothesis test for autocorrelation in the errors in a regression model, the null hypothesis being the absence of autocorrelation. Hence, in terms of normal distribution of errors all regressions except SLR are statistically significant with respect to their Jarque-Bera and Lagrange Multiplier values. Therefore, it can be interpreted that out of these regressions only MLR and Logistic Regression can go for further analysis since they are statistically significant for all parameters.

Table 2. Statistical tests for Regression Analysis

| X | Simple Linear | Multiple Linear | Ridge | LASSO |
|---|---|---|---|---|
| R2/ Pseudo R2 | 0.556 | 0.619 | 0.4855 | 0.46 |
| Adjusted R-Square | 0.556 | 0.618 | 0.48 | 0.4553 |
| F-Statistics | 6007 | 485.0 | 2.449e-37 | 2.477e-27 |
| Durbin-Watson Test | 1.952 | 1.963 | 1.79127 | 1.7163 |
| Jarque-Bera (JB) Test | 9.617 | 0.91 | 34.778 | 90.18286 |
| Lagrange Multiplier Statistic | 14.237 | 145.0 | 208.301 | 185.15 |
| Accuracy | 0.71 | 0.711 | 0.72 | 0.72 |

## 10.2. Classification

The parameters have been tuned to obtain maximum Silhouette Score which ranges from -1 to 1 and a higher Silhouette Score for a given model indicates better performance within that model. Based on accuracy it is seen that K-Means is not as accurate as SVM.

Table 3. Statistical tests for Classification Analysis

| X | K-Means | SVM |
|---|---|---|
| Silhouette Test | 0.7021 | 0.13387 |
| Accuracy | 0.49934 | 0.71 |

## 10.3. Time Series Analysis

From the autocorrelation and partial autocorrelation plot results it is evident that the AR and MA parameters of the Time Series to be considered should be 1 each. This is supported by the Durbin Watson Statistic being 1.4928194722663908 which indicates positive autocorrelation. The Augmented Dickey Fuller Test was showing stationarity based on the value given. The presence of seasonality was confirmed when the model with seasonality was giving a lower AIC value. The presence of exogenous variables was confirmed when they resulted in a reduction in the RMS value. The model was tuned to get the lowest possible Likelihood, AIC, BIC and HQIC. The Ljung Box Statistic leading to a p-value greater than the significance level also shows the validity of the model. The Jarque-Bera Statistic confirms the heteroscedasticity.

Table 4. Time Series Analysis Statistical Test Observations

| X | Older movies analysis |
|---|---|
| Autocorrelation | Cuts of to 0 after 1 lag |
| Partial Autocorrelation | Cuts of to 0 after 1 lag |
| Augmented Dickey-Fuller Test Statistic | -10.462528698062133 |
| RMSE | 62.19 |
| Log-Likelihood | -965.539 |
| AIC | 1949.079 |
| BIC | 1982.679 |
| HQIC | 1962.512 |
| Ljung-Box (Q) | 24.64 |
| Skewness | -0.26 |
| Kurtosis | 4.13 |
| Jarque-Bera (JB) Test | 19.9 |
| Heteroscedasticity | 0.95 |
| Durbin-Watson Test | 1.4928194722663908 |

## 10.4. Artificial Neural Network

The neural network shows its superiority by giving a very high accuracy with the minimum possible loss. The tuned hyper-parameters are given below. The model does not show overfitting.

Table 5. Artificial Neural Network

| Attributes | 'duration', 'avg_vote', 'Action', 'Adventure', 'Animation', 'Biography', 'Comedy', 'Crime', 'Drama', 'Family', 'Fantasy', 'Horror', 'Mystery', 'Thriller' |
|---|---|
| Type | Multi-layer Perceptron Classifier |
| Architecture | İnput Layer: 14<br>Hidden Layer: 100<br>Output Layer: 3 |
| Output Type | Ternary |
| Initial Loss | 0.6915657421532461 |
| Final Loss | 0.1624720046108523 |
| Activation Function | Logistic |
| Optimizer | Adam |
| Early Stopping | True |
| Validation Fraction | 0.1 |
| Number of training examples | 4796 |
| Loss Curve Type | Strictly Decreasing |
| Testing results with 2020 dataset | 93.055555556 % |

## 10.5. Comparison of Valid Models with Similar Accuracy

SLR and Ridge and Lasso Regression were not considered as they proved to be invalid by the normality test and F Statistic respectively. Now 3 models of comparable accuracy exist as shown in Table 5. Hence, a Jaccard Index is used to find the similarity between the attributes obtained from the dataset and that scraped from the IMDb website. The Jaccard Index is used as an indication of the degree to which attributes used in the model are available. From this it can be concluded that SVM is the better model in terms of availability among the 3 followed by Multiple Linear Regression and Logistic Regression respectively.

Table 6. Comparison of Models with Similar Accuracy

| Model Name | Attributes | Attributes 2020 | Jaccard Index |
|---|---|---|---|
| Multiple Linear regression | 'budget','reviews_from_users','review_from_critics', 'top1000_voters_ratings', 'Action','Animation','Crime', 'Drama','Family','Fantasy', 'Horror', 'Music', 'Musical', 'Mystery','Sport','Thriller' | 'duration','Action','Animation','Biography','Drama', 'Horror' | 4/18=0.222 |
| Logistic regression | 'top1000_voters_rating', 'Action','Crime','Drama', 'Fantasy','Mystery','Romance', 'Sport', 'Thriller,' 'War' | 'avg_vote','Action','Crime', 'Fantasy','Mystery' | 4/11=0.18 |
| SVM | 'top1000_voters_rating', 'Action', 'Crime', 'Drama', 'Fantasy', 'Mystery','Romance', 'Sport','Thriller', 'War' | 'avg_vote', 'Action','Crime', 'Drama','Fantasy','Mystery',' Thriller' | 6/11=0.545 |

## 10.6. Some Movie Success Prediction Examples

Table 7 shows all the model results for the most recent movies in the testing dataset used. Here H stands for Hit, F stands for Flop and N stands for Neutral.

Table 7. Movie Success Prediction Examples

| Movie Name | True Success | SLR | MLR | KMeans | Logistic | Ridge | Lasso | SVM | ANN |
|---|---|---|---|---|---|---|---|---|---|
| Jeanne | F | N | N | N | N | N | N | N | N |
| I Trapped the Devil | N | N | N | N | F | N | N | N | H |
| Midsommar | H | N | N | N | F | N | N | N | H |
| Knives Out | H | H | H | N | N | H | H | H | N |
| Sextuplets | F | N | N | N | F | N | N | N | H |
| Unplanned | F | N | F | N | F | F | F | F | F |
| Cold Blood Legacy | F | N | F | N | F | N | F | N | F |
| Playing with Fire | F | H | F | N | F | F | F | F | H |
| Jexi | N | N | N | N | F | N | F | N | H |
| Tomasso | N | N | N | N | F | N | N | N | N |

## 11. Discussion

Our analysis on the prediction of a movie success using traditional regression methods was inspired by the fact that there were plenty of papers that used various machine learning models to predict the success of a movie. Given that these returned accurate results, we wanted to explore these techniques and the statistical significance of using such models. The tests were aimed at checking for the basic assumptions that follow the application of a model such as the normal distribution of errors for regression and so on. The reason for doing so was that we did not want any inconsistency or overfitting to affect our predictions. The end goal was to make the predictor useful in the real world. To corroborate this, the most recent 2020 data was scraped. This gave a sense of the attributes that will be available on immediate release of a movie.

## 12. Conclusion and Future Work

The prediction values from each of the models thus varies depending on their structural build as well as the method they use for prediction, and analysing these changes through comparison as well as statistical tests to prove its significance. By doing so it is seen that the Artificial Neural Network is the best model for prediction followed by the Support Vector Machine, Multiple Linear Regression and Logistic Regression that give comparable performance in terms of accuracy Logistic Regression being slightly higher. The availability of the attributes of these 3 models have also been analyzed where the Support Vector Machine has better availability. Following these 3 models, comes the K-Means with a much lower accuracy. Simple Linear Regression and the Regularization techniques are deemed invalid due to statistically insignificant results. Further, a Time Series Analysis that uses a SARIMAX model forecasts the metascore value effectively. From the models it is evident that some attributes like the top1000_voters_rating and the genres of the movie play a major role in the prediction of movie success. Also, some genres have more predictability than others is evident from the analysis shown previously. In the near future we aim to look into the various improvements that can be performed on the individual models to boost their performance, and hence broaden our perspective on the classification of a variety of other models with comparable performances to show the significance of each one of them.

## AUTHORS

**Shreya Venugopal**
Student at PES university pursuing B-Tech in Computer Science, member of the ACM student chapter, specialization in Machine Intelligence and Data Science and an avid coder.

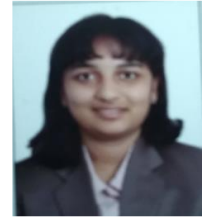

**Rishab Kashyap**
Student at PES university pursuing B-Tech in Computer Science, member of the ACM student chapter, Specialization in Machine Intelligence and Data Science along with minors degree in Electronics and Communication.

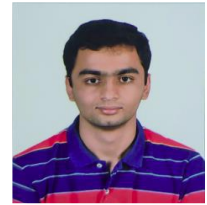

**Manav Agarwal**
Undergraduate Student at PES university pursuing B-Tech in Computer Science, Specialization in Machine Intelligence and Data Science with an innate interest in gadgets and electronics.

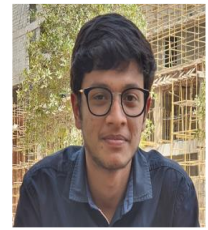

**R Bharathi**
Associate professor at PES University specialisation in Data Science, Machine learning and Data Analytics.

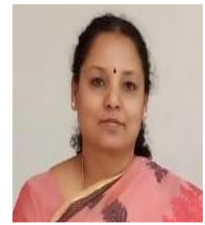